\documentclass[letterpaper, 10 pt, conference]{ieeeconf}  

\IEEEoverridecommandlockouts                              

\usepackage{float} 
\usepackage{subfigure}
\usepackage{booktabs} 
\usepackage{amssymb}
\usepackage{cite}
\usepackage{amsmath}
\usepackage{threeparttable}
\usepackage{multirow}
\usepackage{graphicx}
\usepackage{bm}
\usepackage{multirow}
\usepackage{algorithm}
\usepackage{algpseudocode}
\usepackage{amsmath}
\usepackage{hyperref}

\title{\LARGE \bf
PL-VINS: Real-Time Monocular Visual-Inertial SLAM with Point and Line Features
}

\author{Qiang Fu$^{12*}$, Jialong Wang$^{1*}$, Hongshan Yu$^{1}$, Islam Ali$^{2}$, Feng Guo$^{1}$, \\ Yijia He$^{3}$, Hong Zhang$^{2}$,~\IEEEmembership{Fellow,~IEEE} 
\thanks{This work was supported in part by the National Natural Science Foundation of China under (Grants 61973106, U1813205, and U1913202), the China Scholarship Council under (Grant 201906130082), the Key Research and Development Project of Science and Technology Plan of Hunan Province (Grant 2018GK2021), and the Key Project of Science and Technology Plan of Changsha City (Grant kq1801003). 
\textit{Corresponding author: Hongshan Yu}
}
\thanks{$^{1}$The authors are with National Engineering Laboratory for Robot Visual Perception and Control Technology, Hunan University, China. Email: {\tt\small cn.fq@qq.com}}%
\thanks{$^{2}$The authors are with Department of Computing Science, University of Alberta, Canada. Email: {\tt\small hzhang@ualberta.ca}}%
\thanks{$^{3}$The authors are with Megvii Technology,  Beijing, China. Email: {\tt\small heyijia2016@gmail.com}}%
\thanks{* equal contribution. Demo: bilibili.com/video/BV1464y1F7hk}
}

\begin{document}

\maketitle
\thispagestyle{empty}
\pagestyle{empty}

\begin{abstract}

Leveraging line features to improve localization accuracy of point-based visual-inertial SLAM (VINS) is gaining interest as they provide additional constraints on scene structure. However, real-time performance when incorporating line features in VINS has not been addressed. This paper presents PL-VINS, a real-time and high-efficiency optimization-based monocular VINS method with point and line features, developed based on the state-of-the-art point-based VINS-Mono \cite{vins}. We observe that current works use the LSD \cite{lsd} algorithm to extract line features; however, LSD is designed for scene shape representation instead of the pose estimation problem, which becomes the bottleneck for the real-time performance due to its high computational cost. In this paper, we modify the LSD algorithm by studying a hidden parameter tuning and length rejection strategy. The modified LSD algorithm can run at least three times as fast as LSD. Further, by representing space lines with the Pl\"{u}cker coordinates, the residual error in line estimation is modeled in terms of the point-to-line distance, which is then minimized by iteratively updating the minimum four-parameter orthonormal representation of the Pl\"{u}cker coordinates. Experiments in a public benchmark dataset show that the localization error of our method is 12-16\% less than that of VINS-Mono at the same pose update frequency. 
The source code of our method is available at: https://github.com/cnqiangfu/PL-VINS.
\end{abstract}

\section{INTRODUCTION}
Real-time high-accuracy 6 degree of freedom (DoF) pose estimation in a challenging scene is crucial for many applications \cite{vins, lsd, fu2,fu1,wen, cnnsvo, plslam, looshing, fastorbslam}, such as robotic navigation, unmanned aerial vehicles, and augmented reality. A monocular visual-inertial SLAM method provides a solution to the pose estimation problem with the minimal sensor set of one camera and one inertial measurement unit (IMU), and it has an obvious advantage in terms of weight and cost \cite{strifovio, stuctvio, orbslam3}.

\begin{figure}[t]
\centering  
	\subfigure[VINS-Mono]{
	\includegraphics[width=0.155\textwidth]{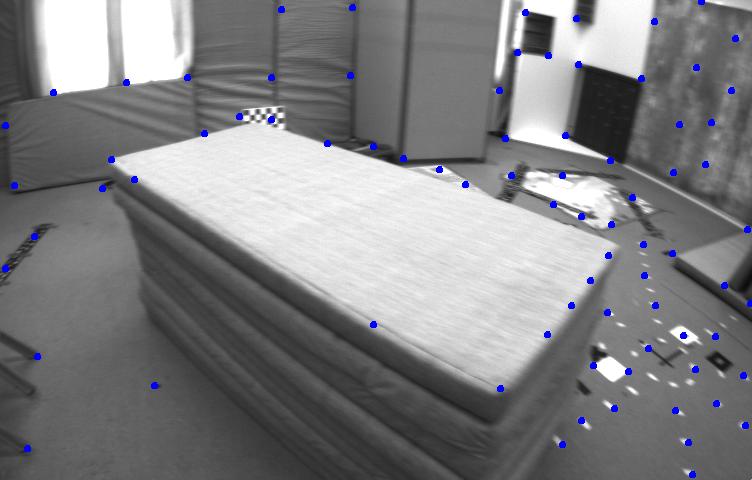}
	\includegraphics[width=0.155\textwidth]{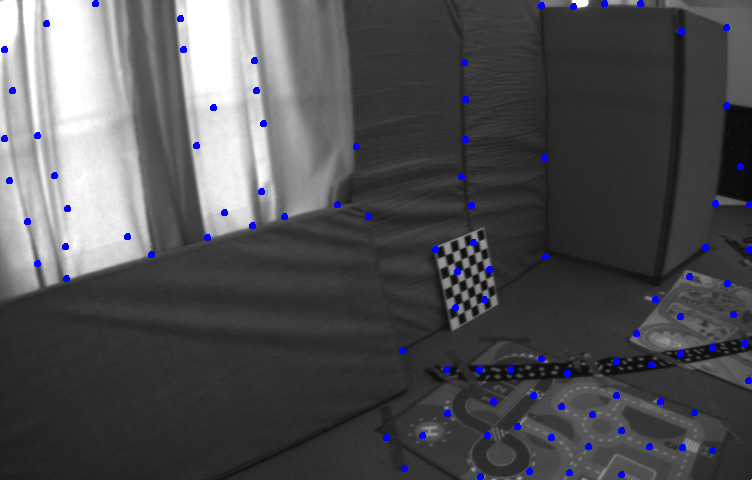}
	\includegraphics[width=0.155\textwidth]{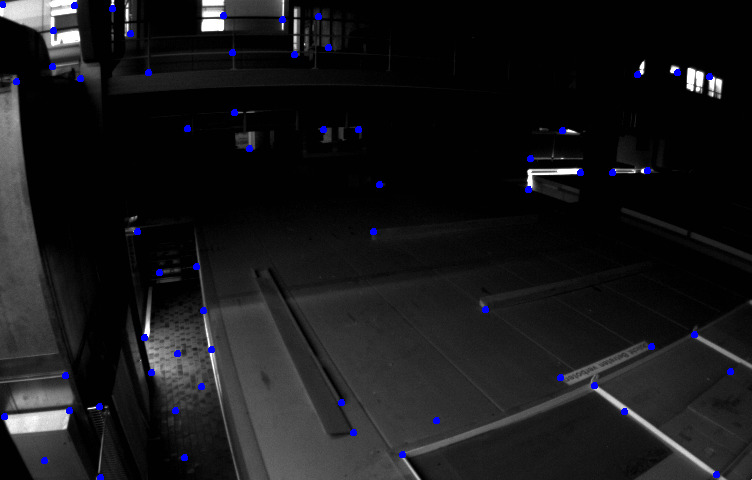}
	}
	\quad
	\subfigure[PL-VINS]{
	\includegraphics[width=0.155\textwidth]{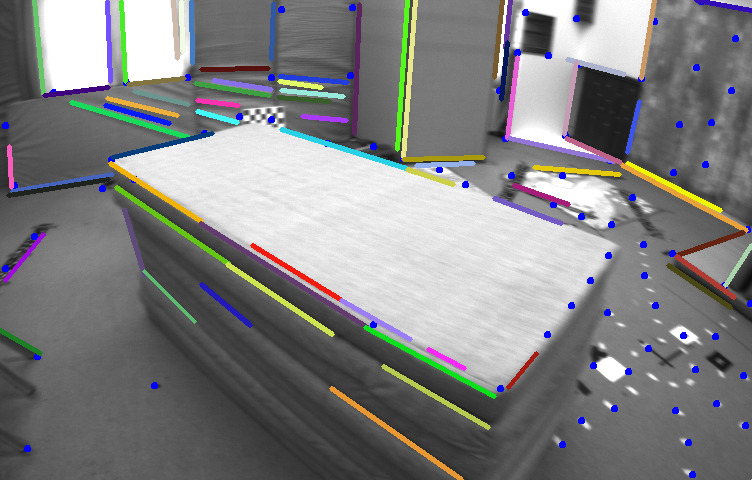}
	\includegraphics[width=0.155\textwidth]{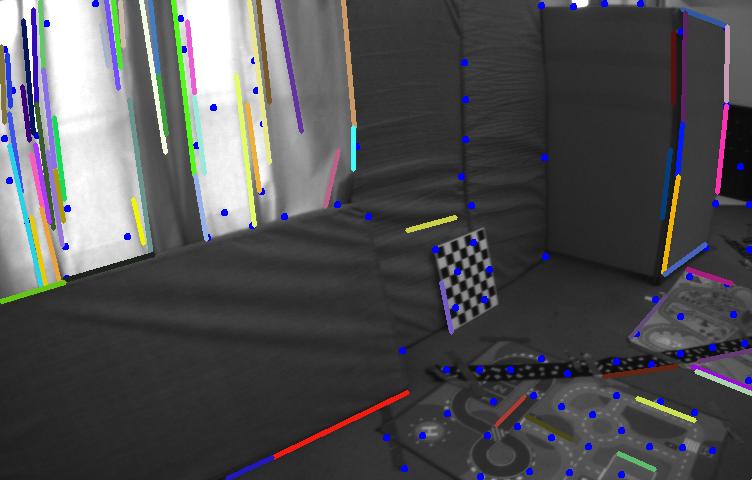}
	\includegraphics[width=0.155\textwidth]{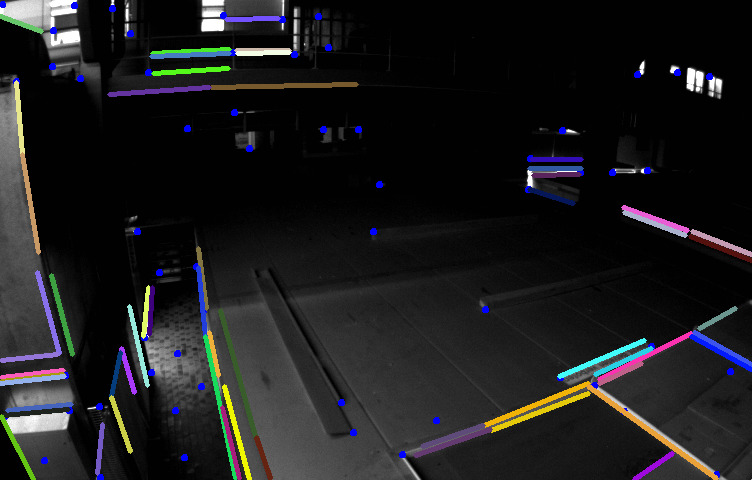}
	}			
\caption{Feature extraction comparison of VINS-Mono and PL-VINS. Images are from EuRoc dataset \cite{euroc}. VINS-Mono uses the Shi-Tomasi algorithm \cite{shi} to extract point features (maximum = 150). However, in these challenging scenes, VINS-Mono may yield a low-accuracy pose estimation due to its dependency on point extraction, whereas PL-VINS leverages the additional line features that are extracted by the modified LSD algorithm.}\label{f:1}
\end{figure}

In the last several years, many monocular VINS methods have been proposed to recover robot motion by tracking point features, such as \cite{okvis, msckf, vins, dsovi1, dsovi2, bloesch, kimera, openvins}. Among them, VINS-Mono is recognized as a non-linear optimization-based benchmark as it yields highly competitive performance with loop closure, pose-graph optimization, and map-merging \cite{orbslam3}. However, we observe it may lead to low-accuracy or failure in pose estimation due to its dependency on point extraction \cite{shi} in some challenging scenes, as Fig. \ref{f:1} shows. Lately, leveraging line features to improve the performance of point-based VINS is gaining interest because line features provide additional constraints on the scene structure especially in human-made environments, such as \cite{plvio, yangIROS, plssvo, strifovio}. These previous works focus on accuracy and robustness improvement, but neglect the issue of real-time performance. \par

\begin{table*}[h]
\centering
\caption{The Comparison of Recent Several Representative Monocular VINS Methods with Point, Line and Plane}
\label{t:table1}
\setlength{\tabcolsep}{2.9mm}{%
\begin{tabular}{@{}lccccccccccc@{}}
\toprule
\multicolumn{1}{c}{\multirow{2}{*}{Methods}} & \multicolumn{3}{c}{Visual Feature} & \multicolumn{2}{c}{Framework} & \multicolumn{2}{c}{Coupled Mode} & \multirow{2}{*}{Odometry} & \multirow{2}{*}{\begin{tabular}[c]{@{}c@{}}Loop \\ Closure\end{tabular}} & \multirow{2}{*}{\begin{tabular}[c]{@{}c@{}}Real-time \end{tabular}} & \multirow{2}{*}{\begin{tabular}[c]{@{}c@{}}Open \\ Source\end{tabular}} \\ \cmidrule(lr){2-8}
\multicolumn{1}{c}{}                         & Point      & Line      & Plane     & Filter      & Optimization    & Loosely         & Tightly        &                           &                                                                          &                                                                                  &                                                                         \\ \midrule
OpenVINS \cite{openvins}                                    & $\surd$    &           &           & $\surd$     &                 &                 & $\surd$        & $\surd$                   &                                                                          & $\surd$                                                                          & $\surd$                                                                 \\
Yang et al. \cite{yangIROS}                                  & $\surd$    & $\surd$   &           & $\surd$     &                 &                 & $\surd$        & $\surd$                   &                                                                          & --                                                                               &                                                                         \\
Yang et al. \cite{yangtro}                                 & $\surd$    & $\surd$   & $\surd$   & $\surd$     &                 &                 & $\surd$        & $\surd$                   &                                                                          & --                                                                               &                                                                         \\
VINS-Mono \cite{vins}                                   & $\surd$    &           &           &             & $\surd$         &                 & $\surd$        & $\surd$                   & $\surd$                                                                  & $\surd$                                                                          & $\surd$                                                                 \\
Li \textit{et al.} \cite{liplp}                                  & $\surd$    & $\surd$   & $\surd$   &             & $\surd$         &                 & $\surd$        & $\surd$                   &                                                                          &                                                                                  &                                                                         \\
PL-VIO \cite{plvio}                                      & $\surd$    & $\surd$   &           &             & $\surd$         &                 & $\surd$        & $\surd$                   &                                                                          &                                                                                  & $\surd$                                                                 \\
Ours                                          & $\surd$    & $\surd$   &           &             & $\surd$         &                 & $\surd$        & $\surd$                   & $\surd$                                                                  & $\surd$                                                                          & $\surd$                                                                 \\ \bottomrule
\end{tabular}%
}
\begin{flushleft}
The criterion of real time follows VINS-Mono \cite{vins}: if the system can run at 10Hz while providing accurate localization information as output. Li \textit{et al.} and PL-VIO cannot run in real time due to their high computational cost in line feature extraction. ``$-$'' means we cannot test it without an open-source code. To our best knowledge, our method is the first real-time optimization-based monocular VINS method with point and line features.
\end{flushleft} 
\end{table*}

Previous works use the LSD algorithm \cite{lsd} from OpenCV \cite{opencv} to detect line features; however, LSD is designed for scene structure representation, instead of the pose estimation problem. Currently, LSD becomes the bottleneck for real-time performance due to its high computational cost \cite{yangIROS}. As Fig. \ref{f:lsd} shows, we observe that a large number of (over 500) short line features are typically detected by LSD; however, they are difficult to match and may disappear in the next frames. As a result, computation resource to detect, describe and match line features is wasted, 
not to mention the generation of outliers that cannot find matched line features. Based on this observation, we argue that we do not need to 
include all line features in a scene description for the pose estimation problem, but rather we should focus on dominant line features and reject short line features. Toward this end, in this work we modify LSD for the specific problem of pose estimation with VINS-Mono. \par

Typical space line representation using two endpoints \cite{plslam} is inappropriate for the pose estimation problem, as it is difficult to accurately extract and track the endpoints between frames due to viewpoint changes and occlusions \cite{zhang}. In this work, the space lines are represented as infinite lines by the Pl\"{u}cker coordinates, and then the line reprojection residual is modeled in terms of the point-to-line distance, which is then minimized by iteratively updating the minimum four-parameter orthonormal representation of the Pl\"{u}cker coordinates. \par
Finally, by integrating the aforementioned line extraction and representation with VINS-Mono, this paper presents PL-VINS whose features include: 
\begin{itemize}
	\item 
	To our best knowledge, PL-VINS is the first real-time optimization-based monocular VINS method with point and line features.
	\item A modified LSD algorithm is presented for the pose estimation problem by studying a hidden parameter tuning and length rejection strategy. The modified LSD algorithm can run at least three times as fast as LSD.  	
	\item Points, lines, and IMU information are fused efficiently in an optimization-based sliding window for high-accuracy pose estimation. 
	\item Experiments on the benchmark dataset EuRoc \cite{euroc} show that our method yields higher localization accuracy than the state-of-the-art (SOTA) VINS-Mono method at the same pose update frequency.
\end{itemize}
\par
In the remainder of the paper, related work is introduced in Section II, and the architecture of the proposed PL-VINS method is described in Section III. Section IV introduces how we utilize line features in this method. The experiment setup and results are described in Section V. Finally, we provide concluding remarks and describe future works in Section VI.

\begin{figure*}[ht]
\centering  
	\subfigure{
	\includegraphics[width=0.98\textwidth]{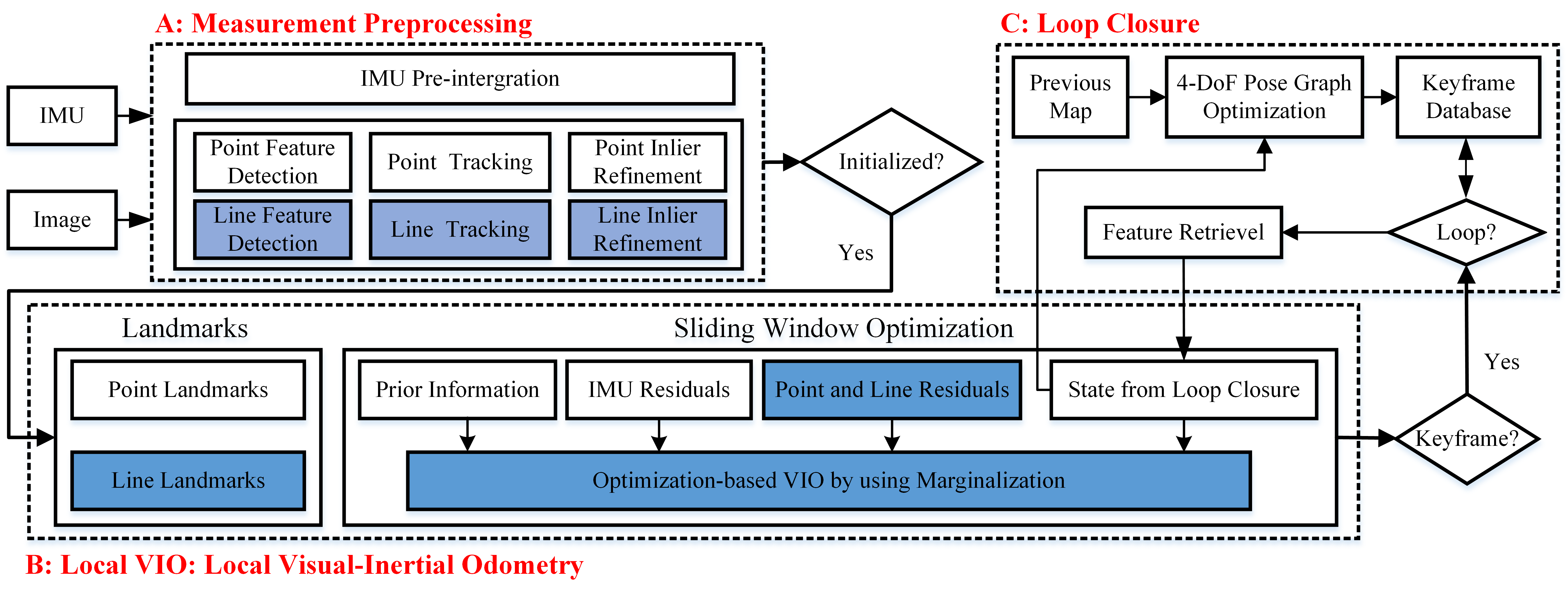}}	
\caption{System overview. PL-VINS is developed based on VINS-Mono and our previous work \cite{fu2}, it implements three threads including \textit{measurement preprocessing}, \textit{local VIO}, and \textit{loop closure}. The Blue rectangles represent the differences from VINS-Mono.}\label{f:system}
\end{figure*}

\section{RELATED WORK}
In terms of the solution type, current VINS methods are filter-based and optimization-based. We review the two groups of VINS methods in this section.\par
\subsection{Filter-based VINS methods}
The first filter-based VINS method is MSCKF\cite{msckf} which is a multi-state constraint Kalman filter (KF) framework. Recently, Patrick \textit{et al.} proposed OpenVINS, an open-source extended KF (EKF)-based platform for visual-inertial estimation. 
After that, Yang \textit{et al.} incorporate line features \cite{yangIROS} or plane features \cite{yangICRA, yangtro} to improve localization accuracy of OpenVINS.

\subsection{Optimization-based VINS methods.}
Compared with filter-based methods, optimization-based VINS methods can jointly optimize two types of measurement residuals: vision (camera) and IMU, to obtain optimal state with the bundle adjustment or graph optimization framework. 
In this paper, the optimization-based methods are of interest to us as they are able to produce accurate localization result despite high computation cost on solving of non-linear optimization. 
Current works are divided into point-based methods and point/line-based methods according to the types of visual features used. \par
\textbf{Point-based methods.}
In the past several years, a number of point-based methods have been proposed by using Shi-Tomasi for point feature extraction, such as \cite{vins, dsovi1, dsovi2, kimera}. Among them, VINS-Mono is recognized as a monocular VINS benchmark as it produces highly competitive localization accuracy; 
however, it may produce low-accuracy pose estimation in a challenging scene due to its dependency on point extraction, as Fig. \ref{f:1} shows. 
\par
\textbf{Point and line-based methods.} Integration of other geometric features into point-based VINS is gaining importance, e.g., line or plane features \cite{liplp}, as they provide additional constraints on the scene structure. Table \ref{t:table1} provides a comparison of recent representative monocular VINS methods with point, line or plane features. Considering real-time performance, point and line-based VINS methods are focused in this work. This group of methods firstly track point and line features concurrently, and pre-integrate the IMU measurement information including gyroscope and accelerometer \cite{vins}. After that, the camera pose is estimated by jointly minimizing three residual residual error terms related to point and line reprojections and IMU \cite{plvio}. \par 
We observe most works when incorporating line features, whether optimization-based methods including PL-VIO \cite{plvio}, \cite{liplp}, and \cite{plssvo}, or filter-based methods including \cite{yangICRA, yangIROS} and \cite{yangtro}, directly use LSD from OpenCV for line feature extraction; however LSD is time-consuming and designed for scene structure representation instead of the pose estimation problem. Currently, the use of LSD has become the bottleneck for real-time performance.

\section{SYSTEM OVERVIEW}
This paper presents PL-VINS, a real-time optimization-based monocular visual-inertial SLAM method with point and line features, in which we efficiently make use of line features to improve the performance of the SOTA point-based VINS-Mono method \cite{vins}. Naturally, the process of the line feature fusion is the focus in this paper. The general structure of PL-VINS is depicted in Fig. \ref{f:system}. \par
Before introducing the structure, we define three necessary coordinate frames: world frame $\pi_{w}$, IMU body frame $\pi_{b}$, and camera frame $\pi_{c}$. The gravity direction is aligned with the $z$-axis of $\pi_{w}$. Camera and IMU motion are considered as of 6 DoFs, and the extrinsic parameters between the IMU and the camera are assumed to be known in advance. PL-VINS implements three threads, to be described in the next three sub-sections. \par

\subsection{Measurement Preprocessing}
PL-VINS starts with this thread whose function is to extract and align two types of raw information measured by camera and IMU, respectively. \par
For each incoming image captured by the camera, point and line features are detected, tracked, and refined concurrently. For the point features, we use Shi-Tomasi \cite{shi} to detect, KLT \cite{klt} to track, and RANSAC-based epipolar geometry constraint \cite{mvg} to identify inliers. For the line features, we modify the original LSD algorithm \cite{lsd} for the pose estimation problem, which can run at least three times as fast as the original LSD. The line features are then described by LBD \cite{lbd} and matched by KNN \cite{opencv}. More details are introduced in Section \ref{s:linemanage}-A and \ref{s:linemanage}-B. For the raw gyroscope and accelerometer measurements by the IMU, we follow VINS-Mono's work to pre-integrate them between two consecutive frames. \par
With the preprocessed information, PL-VINS initializes some necessary values for triggering the next thread. First, a graph structure of up-to-scale poses, space points and lines are estimated in several frames. Next, the graph is aligned with the pre-integrated IMU state. 

\subsection{Local Visual-Inertial Odometry}
After initialization, PL-VINS activates a tightly-coupled optimization-based \textit{local visual-inertial odometry} (VIO) thread for high-accuracy pose estimation by minimizing all measurement residuals, as is described in Section \ref{s:slidingwindow}. \par
First, space points and lines are re-constructed by triangulating the 2D point and line feature correspondences between frames. Space points are parameterized by the inverse depth \cite{inverse}, and space lines are parameterized by the Pl\"{u}cker coordinates, as will be described in Section \ref{s:linelandmark}. \par
Next, a fixed-size sliding window is adopted to find the optimal state vector including pose, velocity, space points and lines, acceleration, and gyroscope bias, as is described by Equation (\ref{e:statevector}), by jointly minimizing several residual functions, defined by Equation (\ref{e:objectfunction}). 
When inputting a new frame, we marginalize the last frame in the sliding window \cite{shen} to maintain the window size. \par

\subsection{Loop Closure}
After \textit{local VIO}, PL-VINS judges whether the current frame is a keyframe, depending upon whether the parallax between current frame and last keyframe is greater than a certain threshold or the number of tracked features is smaller than a certain threshold. Once current frame is selected as a keyframe, PL-VINS activates the \textit{loop closure} thread to search and decide if loop closure has occurred. This thread follows the original VINS-Mono's work.

\section{LINE FEATURE FUSION}
\label{s:linemanage}

\begin{figure}[tp]
\centering
	\subfigure[LSD, 647]{
	\includegraphics[width=0.155\textwidth]{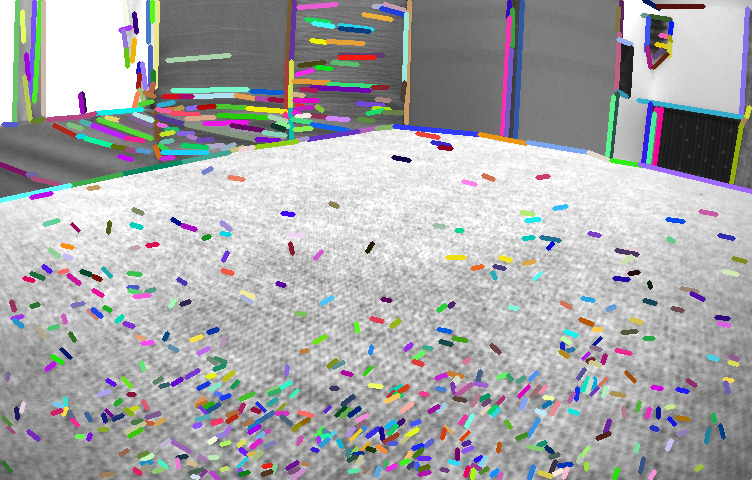}}\hspace{-1mm}
	\subfigure[Length $>$ 20, 107]{
	\includegraphics[width=0.155\textwidth]{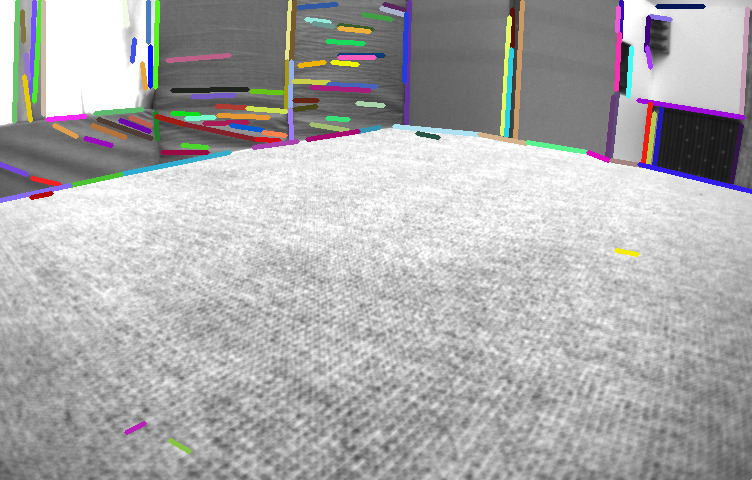}}\hspace{-1mm}	
	\subfigure[Length $>$ 40, 50]{
	\includegraphics[width=0.155\textwidth]{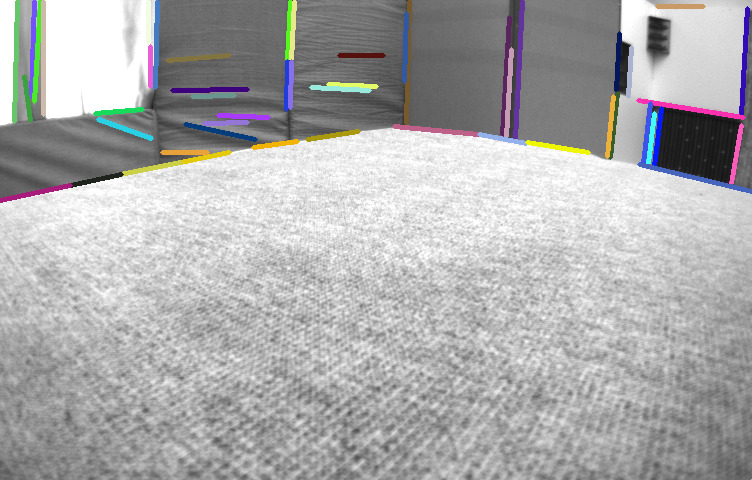}}\hspace{-1mm}				
\caption{LSD is a line segment detector with no parameter tuning and designed for scene shape representation. In this case, it takes 62 ms to detect 647 line features in the image of 782$\times$480 pixels; however, a larger number of short line features can be regarded as outliers for the pose estimation problem as they are difficult to match and may disappear in the next frames, not to mention the difficulty in culling oulters. In the 647 line features, there are 107 line features with length over 20 pixels, and 50 line features with length over 40 pixels. In fact, long line features are easier to match and more likely to appear in next frames than short line features.}\label{f:lsd}
\end{figure}

\subsection{Line Feature Detection}
Current point and line-based VINS methods directly use LSD from OpenCV for line feature extraction, such as \cite{plvio, yangIROS, plssvo, strifovio, yangtro, liplp}; however, LSD has become the bottleneck for the real-time performance due to its high computational cost. We observe that the original LSD algorithm is designed for scene shape representation instead of the pose estimation problem. As Fig. \ref{f:lsd} shows, it takes 62 ms to detect 647 line features in an image of 782$\times$480 pixels; however, a large number of (over 500) short line features are difficult to match and some of them may disappear in the next frames. 
In fact, for the pose estimation problem, it is unnecessary to include all line features of a scene. Instead, we should pay attention to the dominant line features extraction as they are easy to match and likely to appear in the next frames. \par
In this work, we modify the original LSD algorithm for the pose estimation problem based on the source code in OpenCV. This modification process proceeds by first simplifying the line feature detection by studying a hidden parameter tuning, and then filtering out short line features by a length rejection strategy. As a result, the modified LSD algorithm can run three times at least as fast as LSD. 
\par
\textbf{Hidden parameter studies}. LSD is a line segment detector with no parameter tuning \cite{lsd}; however, interestingly, we find there are still several hidden parameters that can be tuned to speed up the line detection for the pose estimation problem. In this work, we make these parameters visible, and all of them can be found in our open-source code. \par
OpenCV uses an $N$-layer Gaussian pyramid to create a multi-scale representation, in which the original image is down-sampled $N$-1 times with a fixed scale ratio $r$ and blurred $N$ times with a Gaussian filter. The above two parameters $r$ and $N$ are available and visible in the original OpenCV function. In this work, we follow a standard setting $r=0.5$ and $N$ = $2$ \cite{plvio, liplp}. After that, OpenCV designs an image scale parameter $s=(0, 1]$ to scale the image of each layer, and then line features are extracted using LSD. 
\par
LSD starts with a line-support region partition operation, in which an image is partitioned into a number of regions where all points share roughly the same image gradient, and each region is approximated with a rectangular. Subsequently, LSD defines a minimal density threshold $d$ to reject line features: the number of aligned points in the rectangle \cite{lsd} needs to be less than the threshold. \par
\begin{figure}[tp]
\centering
	{
	\includegraphics[width=0.23\textwidth]{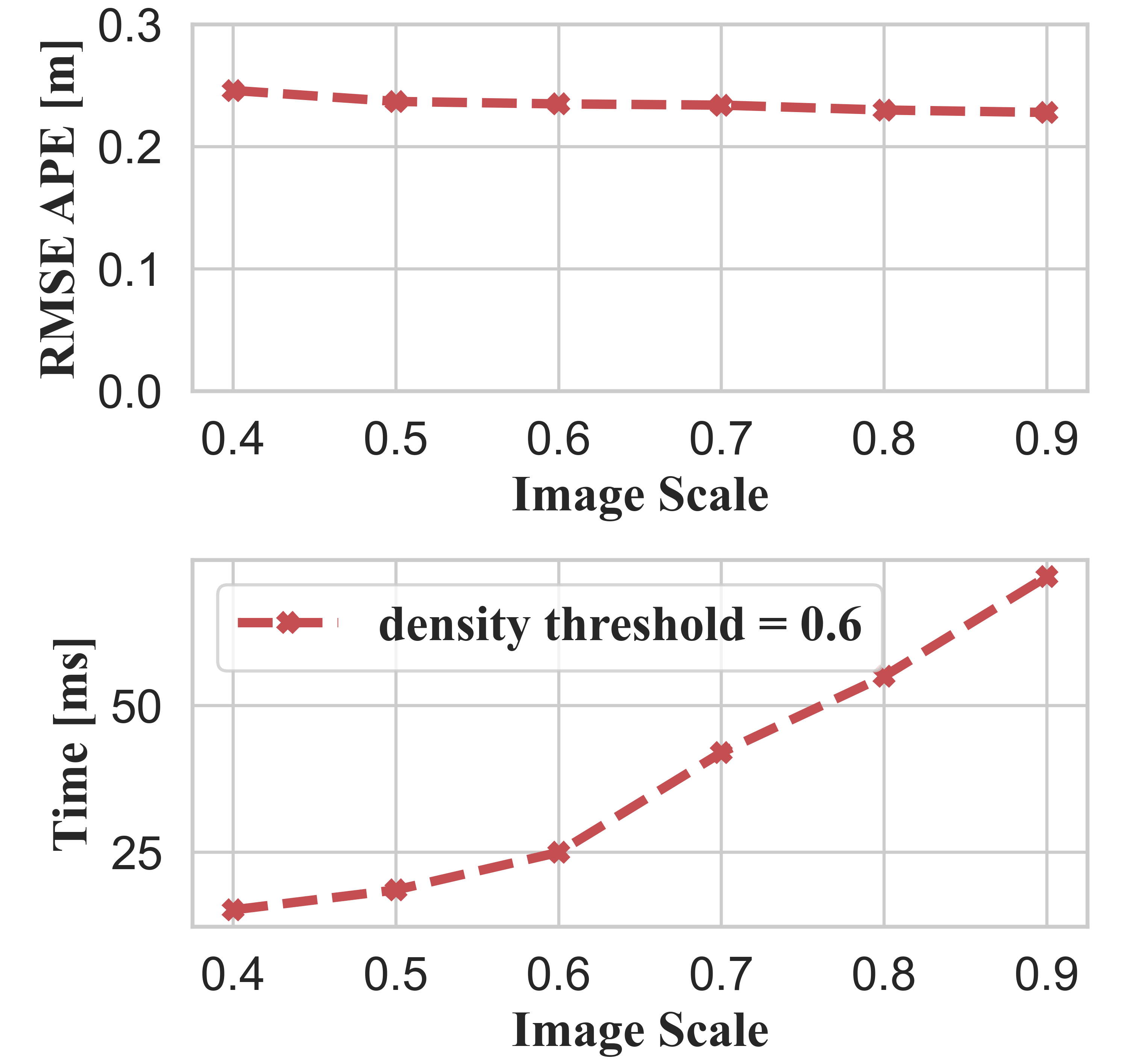}}\hspace{-1mm}
	{
	\includegraphics[width=0.23\textwidth]{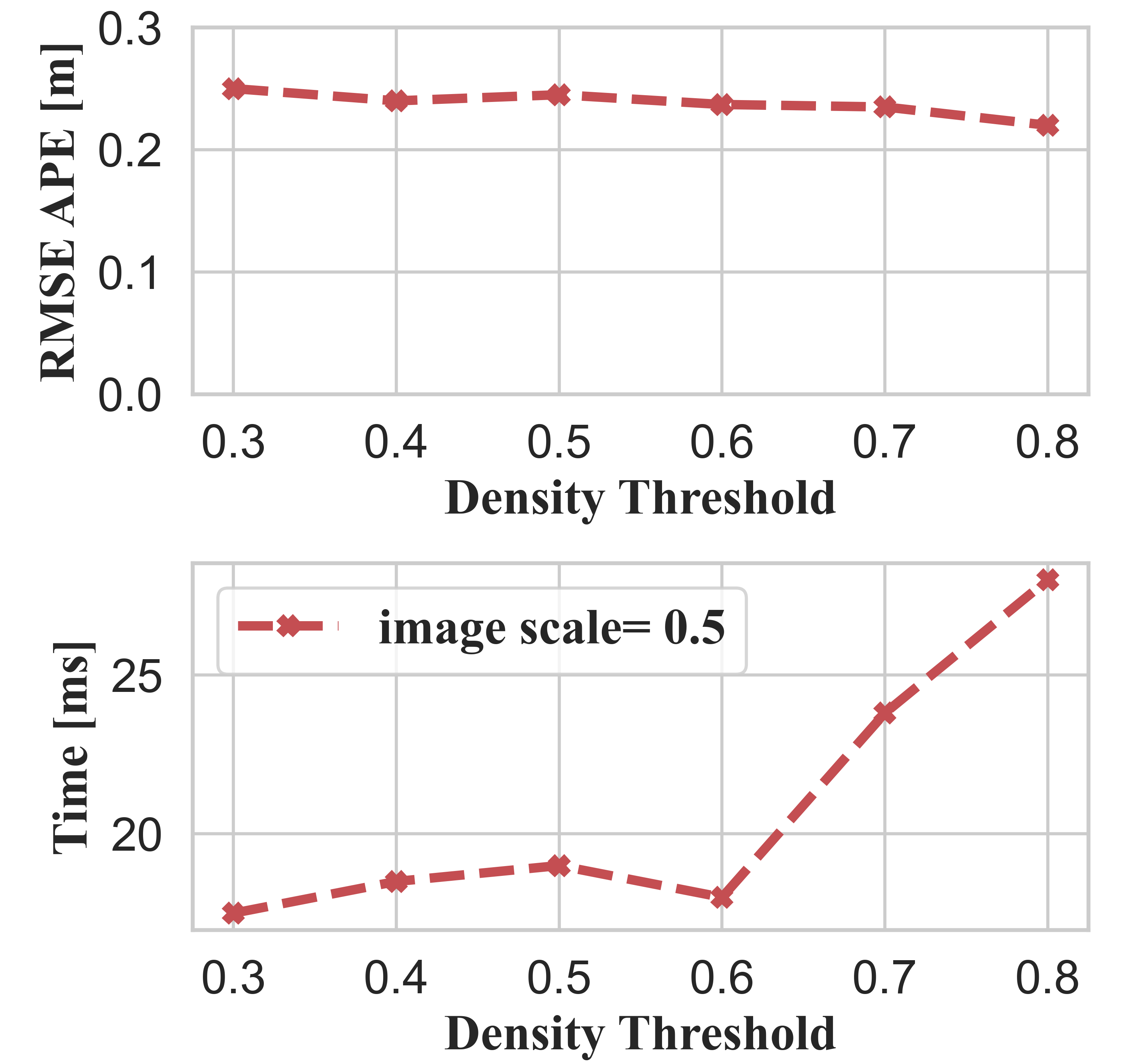}}
	\quad		
\caption{Hidden parameter studies. The results are obtained by running PL-VINS w/o loop on the MH-04-\textit{Difficult} sequence. Time means the average execution time of the modified LSD algorithm. 
Note that decreasing $s$ and $d$ results in significantly time reduction with practically negligible accuracy loss. In this work, we set $s$ = 0.5 (default 0.8) and $d$ = 0.6 (default 0.7) in consideration of efficiency.
}\label{f:parameter}
\end{figure}

Note that we observe that the two hidden parameters $s$ and $d$ are crucial for the time reduction of line feature detection. Fig. \ref{f:parameter} shows a study process. Note that decreasing $s$ and $d$ results in significantly time reduction with practically negligible accuracy loss. In this work, we set $s$ = 0.5 (default 0.8) and  $d$ = 0.6 (default 0.7) in consideration of efficiency and generality. 
As for the rest of the hidden parameters including the angle tolerance, the quantization error tolerance on image gradient norm, and the refinement trigger, we use the default values. \par

\begin{figure}[tp]
\centering	
	\includegraphics[width=0.35\textwidth]{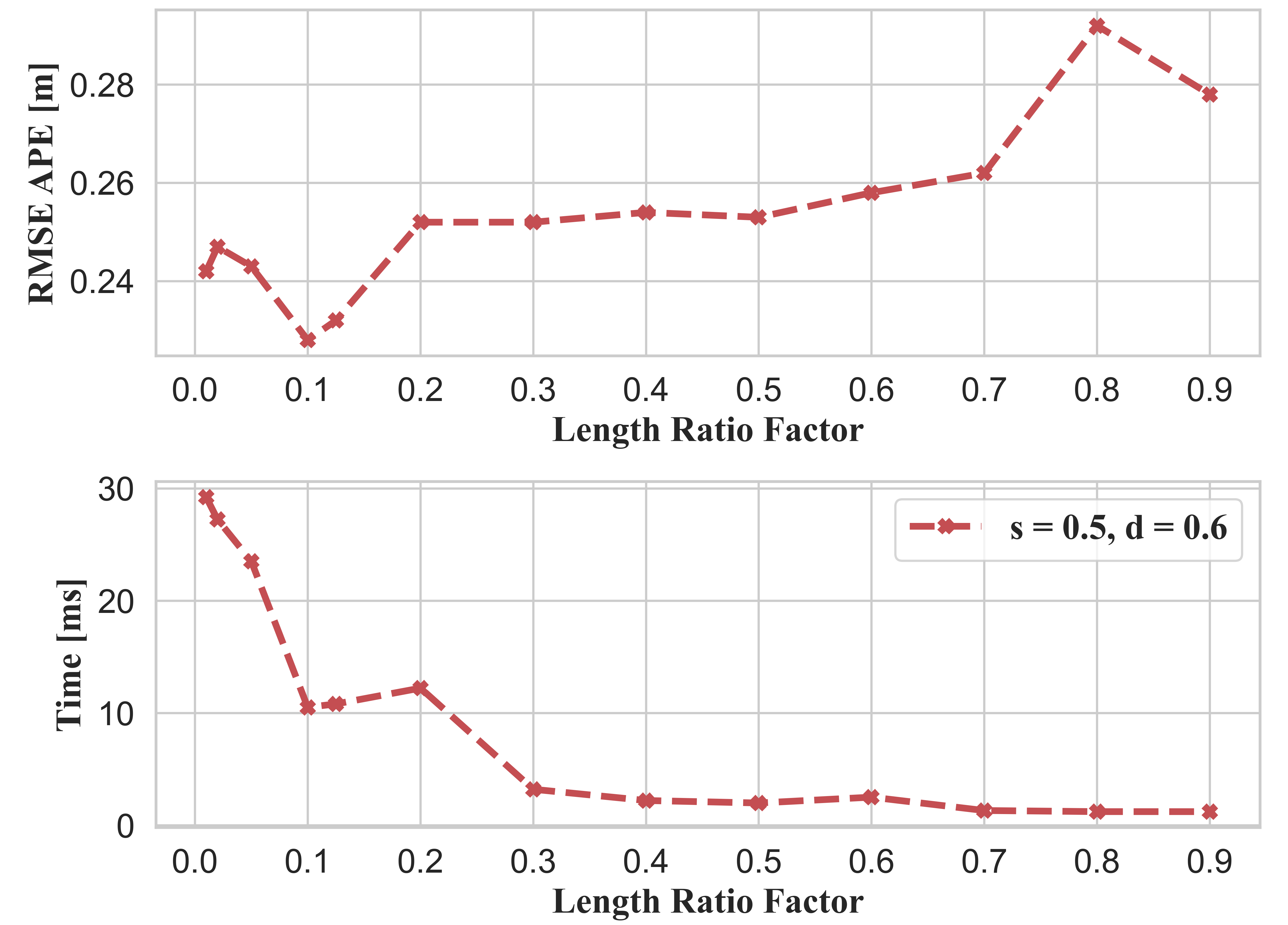}
\caption{Length rejection strategy studies. The results are obtained by running PL-VINS w/o loop on the MH-04-\textit{Difficult} sequence. Time means the average execution time of the line feature tracking. Note that the length ratio factor $\eta$ directly affects the time of the description and match. Higher $\eta$, longer line features will be rejected. We suggest $\eta$ in $[0.1, 0.15]$ in consideration of efficiency. In this work, we set $\eta = 0.125$.
}\label{f:length}
\end{figure}

\textbf{Length rejection strategy}. Previous process simplifies the detection process. Now we use a minimum length threshold $L_{min}$ to reject short line features:
\begin{equation}
L_{min}= \lceil \eta* min(W_{I},H_{I})\rceil,
\end{equation}
where $min(W_{I},H_{I})$ denotes the smaller value between the width and the height of the input image. $\lceil \cdot \rceil$ denotes the ceiling operation. $\eta$ is the ratio factor. Fig. \ref{f:length} shows a study process. In this work we set $\eta = 0.125$ in consideration of generality.
\par

\subsection{Line Tracking and Inlier Refinement}
After line feature detection, we adopt the LBD \cite{lbd} and KNN \cite{opencv} to describe and match line features, respectively. In particular, for possible line outliers, we execute inlier refinement to identify line feature inliers: 
\begin{itemize}
\item the Hamming distance that is computed by KNN needs to be more than 30; 
\item the angular between matching line features needs to be less than 0.1 rad.    
\end{itemize}

\subsection{Space Lines}
\label{s:linelandmark}
In previous steps, we establish line feature correspondences in image plane. Now we initialize space lines (landmarks) in $\pi_{w}$ by triangulating these correspondences. \par
\textbf{Space line representation}. Given a space line $L_{w} \in \pi_{w}$, we describe it with the Pl\"{u}cker coordinates $L_{w}=(\mathbf{n}_{w}^{\top},\mathbf{d}_{w}^{\top})^{\top} \in\mathbb{R}^6$, where $\mathbf{n}_{w} \in \mathbb{R}^3$ denotes the normal vector of the plane determined by $L_{w}$ and the origin of $\pi_{w}$, and $\mathbf{d}_{w} \in \mathbb{R}^3$ denotes the direction vector determined by the two endpoints of $L_{w}$. Then, we initialize the Pl\"{u}cker coordinates. Assume that $L_{w}$ is observed by the two frames whose camera positions are $c_{1}$ and $c_{2}$ in $\pi _{w}$, respectively. Thus we have two planes determined by the two positions and $L_{w}$: $\pi_{1} = (c_{1}, L_{w})$ and $\pi_{2} = (c_{2}, L_{w})$. According to \cite{mvg}, the dual Pl\"{u}cker matrix $L_{w}^{\star}$ can be described as:
\begin{equation}
L_{w}^{\star} = \begin{bmatrix}\left[\mathbf{d}_{w}\right]_{\times} & \mathbf{n}_{w} \\-\mathbf{n}_{w}^{\top} & 0 \end{bmatrix}=\pi_{1}\pi_{2}^{\top}-\pi_{2}\pi_{1}^{\top}. 
\label{e:linetriangulate}
\end{equation} 
where $\left[ \cdot \right]_{\times}$ denotes the skew-symmetric matrix of a 3-vector.
\par
\textbf{Orthonormal representation}. The Pl\"{u}cker coordinates are over-parameterized with a constraint $\mathbf{n}_{w}^{\top} \mathbf{d}_{w}= 0$, so their  dimensions can be reduced further. Zhang \textit{et al.} have verified that the Pl\"{u}cker coordinates can be described with a minimal four-parameter orthonormal representation \cite{zhang}, which has a superior performance in terms of convergence. In this work, we use it for the optimization process. The orthonormal representation $(\mathbf{U}, \mathbf{W}) \in SO(3) \times SO(2)$ of $L_{w}=(\mathbf{n}_{w}^{\top},\mathbf{d}_{w}^{\top})^{\top}$ can be computed using the QR decomposition:
\begin{equation}
\left[\mathbf{n}_{w}\mid \mathbf{d}_{w}\right]=\mathbf{U}\begin{bmatrix}\omega_{1} & 0 \\0 & \omega_{2} \\0&0 \end{bmatrix}, set: \mathbf{W}=\begin{bmatrix}\omega_{1} & \omega_{2} \\-\omega_{2} & \omega_{1} \end{bmatrix},
\label{e:qrcomposition}
\end{equation}
where $\mathbf{U}$ and $\mathbf{W}$ denote a three and a two dimensional rotation matrix, respectively. Let $\mathbf{R}(\bm{\theta})=\mathbf{U}$ and $\mathbf{R}(\theta)=\mathbf{W}$ be the corresponding rotation transformations, we have:
\begin{equation}
\begin{aligned}
\mathbf{R}(\bm{\theta}) & =\left[\mathbf{u}_{1},\mathbf{u}_{2},\mathbf{u}_{3}\right]=\left[\frac{\mathbf{n}_{w}}{||\mathbf{n}_{w}||},\frac{\mathbf{d}_{w}}{||\mathbf{d}_{w}||},\frac{\mathbf{n}_{w}\times\mathbf{d}_{w}}{||\mathbf{n}_{w}\times \mathbf{d}_{w}||} \right] \\
\mathbf{R}(\theta) & =\begin{bmatrix}\omega_{1} & \omega_{2} \\-\omega_{2} & \omega_{1} \end{bmatrix}=\begin{bmatrix}\cos(\theta) & -\sin(\theta) \\\sin(\theta) & \cos(\theta) \end{bmatrix} \\ 
& =\frac{1}{\sqrt{(||\mathbf{n}_{w}||^{2}+||\mathbf{d}_{w}||^{2})}}\begin{bmatrix}||\mathbf{n}_{w}|| & -||\mathbf{d}_{w}||\\||\mathbf{d}_{w}|| & ||\mathbf{n}_{w}|| \end{bmatrix}
\end{aligned},
\label{e:rotation1}
\end{equation}
where $\bm{\theta}$ and $\theta$ denote a 3-vector and a scalar, respectively. Now we define the orthonormal representation by a four-parameter vector:
\begin{equation} 
\mathbf{o}^{\top}= (\bm{\theta}^{\top},\theta).
\label{e:orthonormal}
\end{equation}
Similarly, given an orthonormal representation $(\mathbf{U}, \mathbf{W})$, we can recover its Pl\"{u}cker coordinates by 
\begin{equation}
\mathbf{L}_{w}=\left[\omega_{1}\mathbf{u}_{1}^{\top}, \omega_{2}\mathbf{u}_{2}^{\top}\right],
\end{equation}
where $\omega_{1}$, $\omega_{2}$, $\mathbf{u}_{1}$, and $\mathbf{u}_{2}$ can be extracted from Equation (\ref{e:rotation1}). Note that the orthonormal representation is only used for the optimization process, as is described in Section \ref{s:slidingwindow}. 
\subsection{Line Reprojection Residual Model}
The line reprojection residual is modeled in terms of point-to-line distance. First, we define line geometry transformation. Given a transformation matrix $\mathbf{T}_{cw} = \begin{bmatrix}\mathbf{R}_{cw}, \mathbf{t}_{cw} \end{bmatrix}$ from $\pi_{w}$ to $\pi_{c}$, where $\mathbf{R}_{cw} \in SO(3)$ and $\mathbf{t}_{cw} \in \mathbb{R}^{3}$ define rotation and translation, respectively. With the matrix, we can transform $L_{w}$ in $\pi_{w}$ to $\pi_{c}$ by:
\begin{equation}
L_{c}=\begin{bmatrix}\mathbf{n}_{c} \\ \mathbf{d}_{c} \end{bmatrix} = \mathbf{T}_{cw}L_{w}=\begin{bmatrix}\mathbf{R}_{cw} & \left[\mathbf{t}_{cw}\right]_{\times}\mathbf{R}_{cw} \\0 & \mathbf{R}_{cw} \end{bmatrix}\begin{bmatrix}\mathbf{n}_{w} \\ \mathbf{d}_{w} \end{bmatrix},
\label{e:linetrans}
\end{equation} 
where $L_{c}$ is the Pl\"{u}cker coordinates of $L_{w}$ in $\pi_{c}$. \par 
Next, the projection line $\mathbf{l}$ is obtained by transforming $L_{c}$ to the image plane \cite{zhang}:
\begin{equation}
\mathbf{l} = [l_{1},l_{2},l_{3}]^{\top} = \mathbf{K}_{L}\mathbf{n}_{c},
\end{equation}
where $\mathbf{K}_{L}$ denotes the line projection matrix and $\mathbf{n}_{c}$ can be extracted by Equation (\ref{e:linetrans}). \par
Finally, assume that $L_{w}$ denotes the $j$-th space line $\mathcal{L}_{j}$ which is observed by the $i$-th camera frame $c_{i}$. The line re-projection error can be defined as:
\begin{equation}
\mathbf{r}_{L}\left(\mathbf{z}_{\mathcal{L}_{j}}^{c_{i}}, \mathcal{X}\right) = \mathbf{d}(\mathbf{m},\mathbf{l}) = \frac{\underline{\mathbf{m}}^{\top}\mathbf{l}}{\sqrt{l_{1}^{2}+l_{2}^{2}}} \in \mathbb{R}^{1}
\label{e:linereprojecterror}
\end{equation}
where $\mathbf{d}(\mathbf{m},\mathbf{l})$ denotes the point-to-line distance function. $\underline{\mathbf{m}}$ is the homogeneous coordinates of the midpoint of a line feature. The corresponding Jacobian matrix $\mathbf{J}_{L}$ can be obtained by the chain rule \cite{plvio}. 
\par


\begin{figure*}[t]
\flushleft  
	\subfigure[VINS-Mono w/o loop]{
	\includegraphics[width=0.25\textwidth]{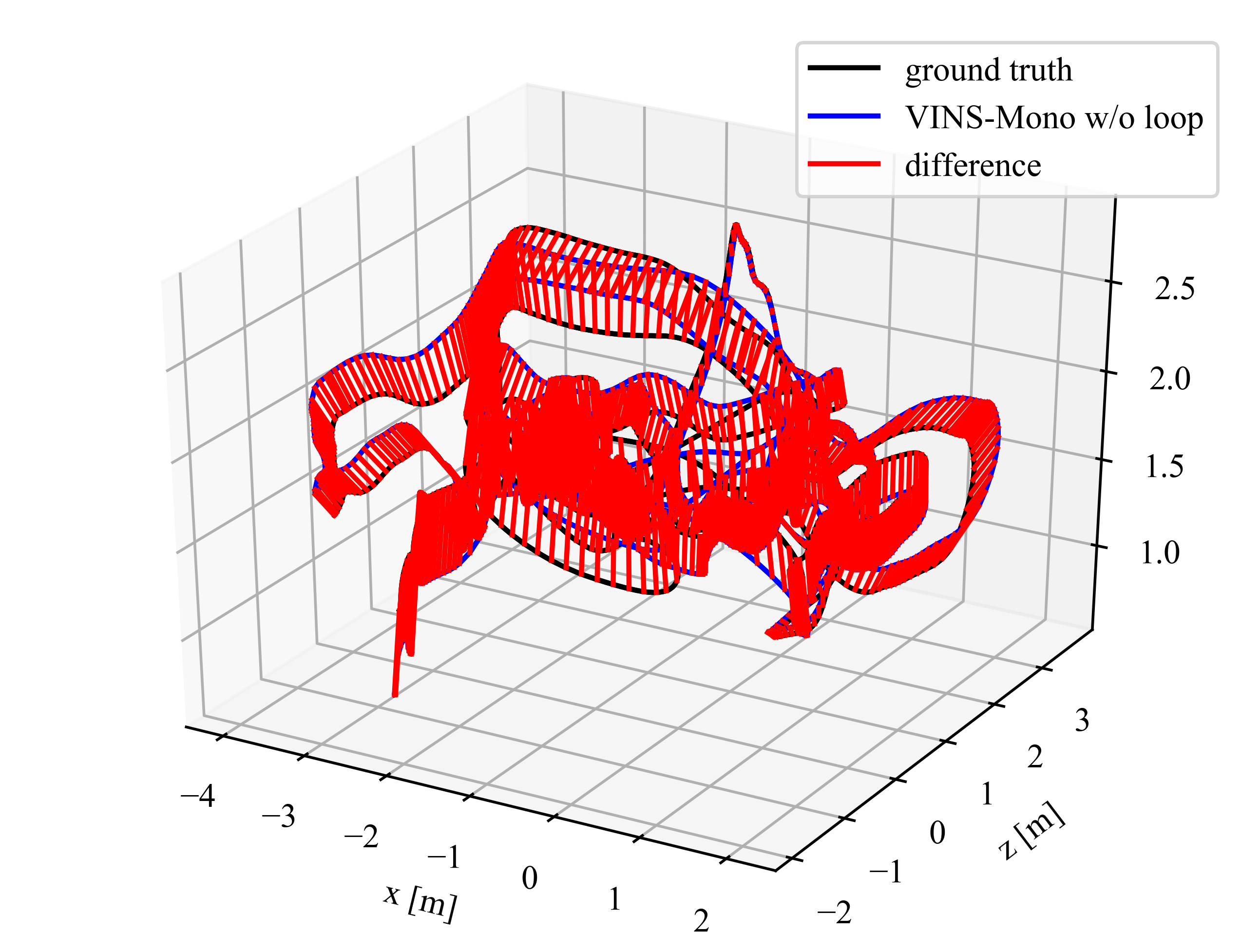}}\hspace{-3mm}
	\subfigure[PL-VINS w/o loop]{
	\includegraphics[width=0.25\textwidth]{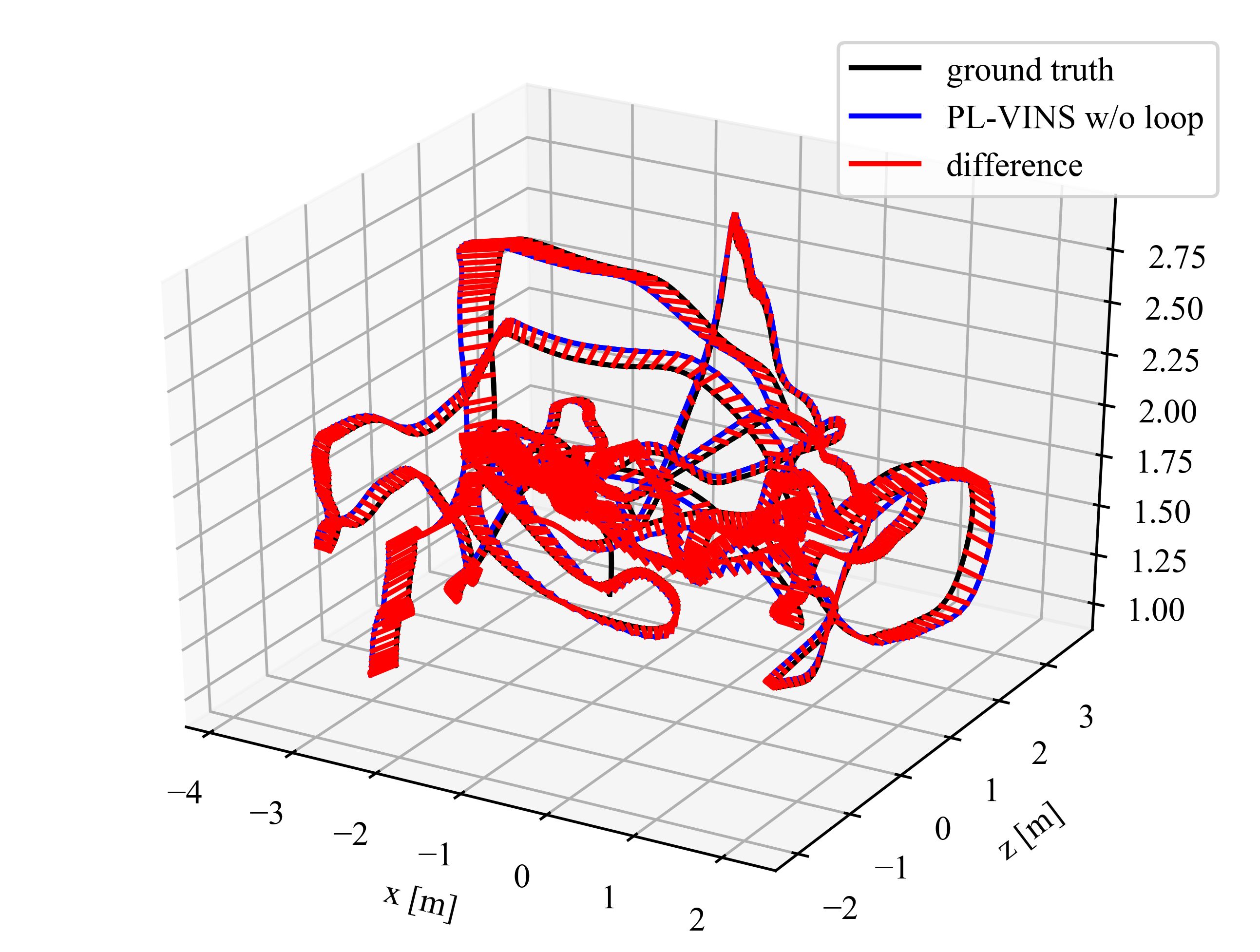}}\hspace{-3mm}
	\subfigure[VINS-Mono]{
	\includegraphics[width=0.25\textwidth]{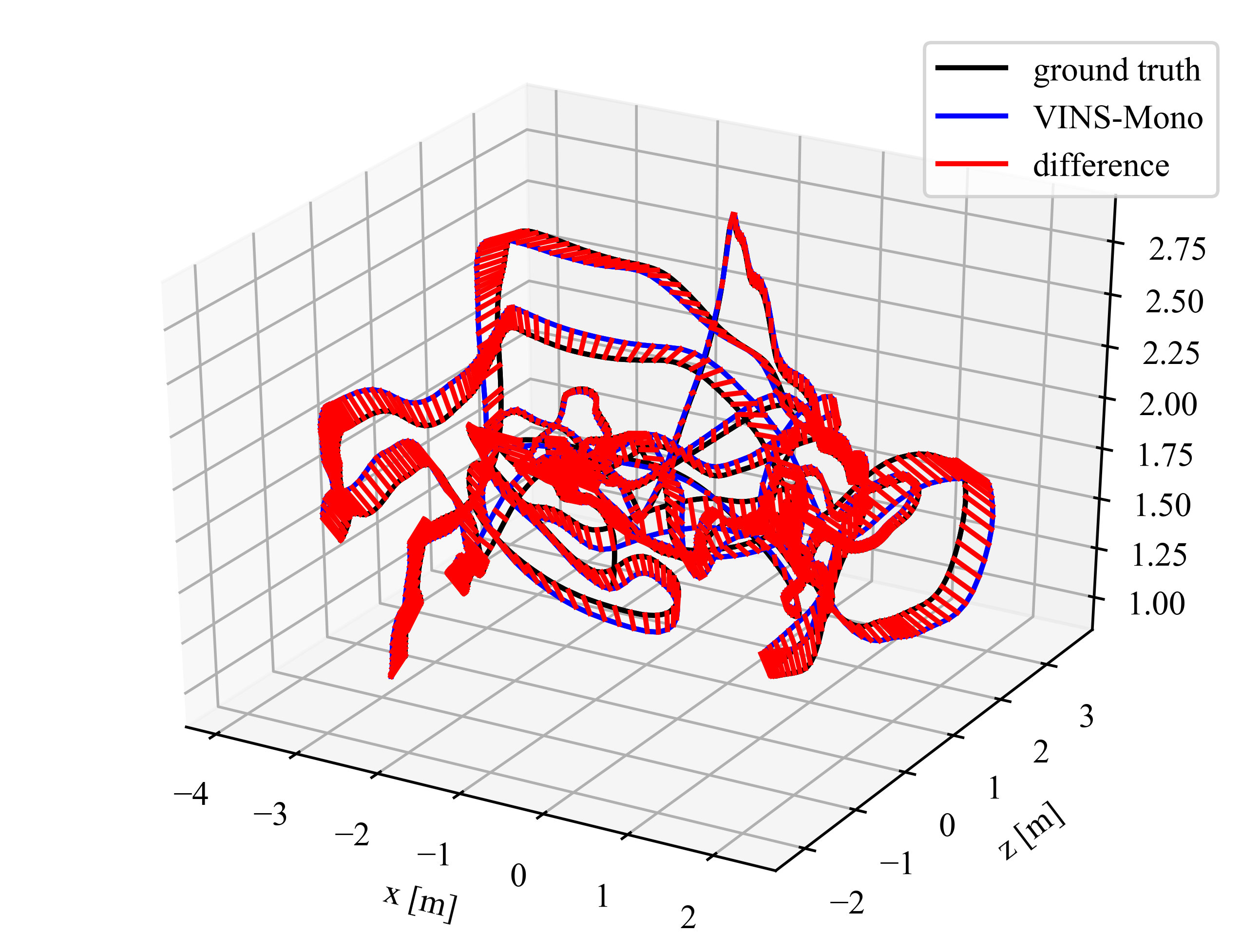}}\hspace{-3mm}
	\subfigure[PL-VINS]{
	\includegraphics[width=0.25\textwidth]{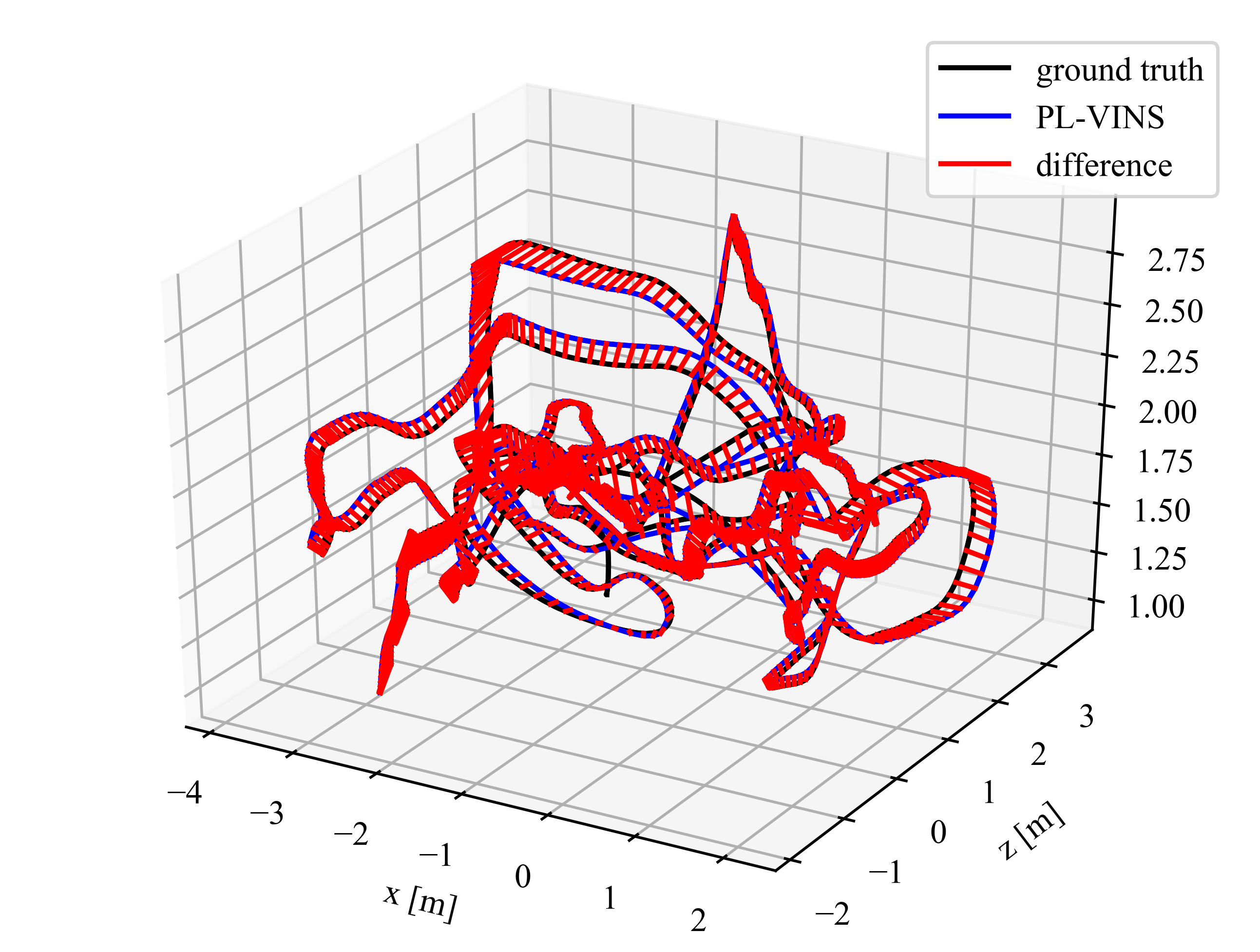}}\hspace{-3mm}	
\caption{3D trajectory comparision of VINS-Mono and PL-VINS w/ and w/o loop on the V2-03-\textit{difficult} sequence. The red line denotes localization error. Quantitative results can be found in Table \ref{t:ate}. It can be seen that PL-VINS produces better localization accuracy whether w/ or w/o loop.} \label{f:3dtrajectory}
\end{figure*}

\begin{figure}[t]
\centering  
	\subfigure[VINS-Mono]{	
	\includegraphics[width=0.235\textwidth]{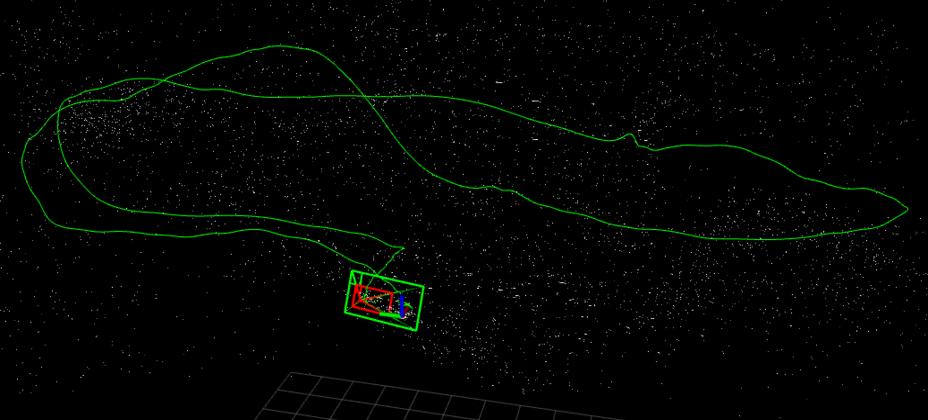}}\hspace{-1mm}
	\subfigure[PL-VINS]{
	\includegraphics[width=0.235\textwidth]{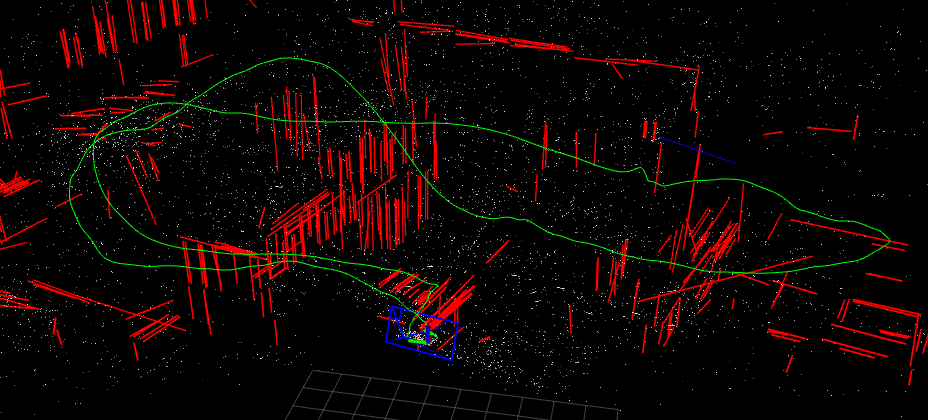}}
\caption{Trajectory and space feature map comparison of VINS-Mono and PL-VINS on the MH-04-\textit{Difficult} sequence. The two pictures are the screenshots of ROS Rviz window, where the yellow line denotes the motion trajectory, the white points represent space points, and the orange lines represent space lines. Quantitative results can be found in Table \ref{t:ate}.}\label{f:ros}
\end{figure}

\subsection{Sliding Window Optimization with Point, Line and IMU}
\label{s:slidingwindow}
First, we define the full state vector $\mathcal{X}$ in the sliding window related to point, line and IMU measurement information:
\begin{equation}
\begin{aligned}
& \mathcal{X}= \left[\mathbf{x}_{0}+\mathbf{x}_{1},...,\mathbf{x}_{n_{k}}, \lambda_{0},\lambda_{1},...\lambda_{n_{p}}, \mathbf{o}_{1}, \mathbf{o}_{2},...,\mathbf{o}_{n_{l}}\right] \\
& \mathbf{x}_{k}=\left[\mathbf{p}_{b_{k}^{w}},\mathbf{q}_{b_{k}^{w}},\mathbf{v}_{b_{k}^{w}},\mathbf{b}_{a},\mathbf{b}_{g} \right], k \in \left[0,n_{k} \right]
\end{aligned}
\label{e:statevector}
\end{equation}
where $\mathbf{x}_{k}$ contains the $k$-th IMU body position $\mathbf{p}_{b_{k}^{w}}$, orientation $\mathbf{q}_{b_{k}^{w}}$, velocity $\mathbf{v}_{b_{k}^{w}}$ in $\pi_{w}$, acceleration bias $\mathbf{b}_{a}$, gyroscope bias $\mathbf{b}_{g}$. $n_{k}$, $n_{p}$, and $n_{l}$ denote the total number of keyframes, space points and lines in the sliding window, respectively. $\lambda$ is the inverse distance of a point feature from its first observed keyframe. $\mathbf{o}^{\top}= (\bm{\theta}^{\top},\theta)$ is the four-parameter orthonormal representation of a 3D line, as is described by (\ref{e:orthonormal}). 
\par
Next, we define the objective function in the sliding window:
\begin{equation}
\mathop{\min}_{\mathcal{X}}
\left( \mathbf{I}_{prior}+\mathbf{e}_{imu}+ \mathbf{e}_{point}+ \mathbf{e}_{line}+\mathbf{e}_{loop}\right),
\label{e:objectfunction}
\end{equation}
where $\mathbf{e}_{prior}$ denotes the prior information, which is obtained when marginalizing out an old frame in the sliding window \cite{shen}. $\mathbf{e}_{imu}$ and $ \mathbf{e}_{point}$ denote point reprojection residual and IMU measurement residual, respectively. $\mathbf{e}_{loop}$ denotes the loop-closure constraint. The aforementioned four terms have been introduced in VINS-Mono. \par
Compare with VINS-Mono, this work modifies its object function with an additional line reprojection residual term $\mathbf{e}_{line}$, which we will define. Let $\mathcal{L}$ be the set of line features observed in the sliding window, and $(i,j) \in \mathcal{L}$ denotes that the $j$-th space line $\mathcal{L}_{j}$ is observed by the $i$-th camera frame $c_{i}$ in the sliding windows. Based on Equation (\ref{e:linereprojecterror}), $\mathbf{e}_{line}$ can be defined by
\begin{equation}
\begin{cases} \mathbf{e}_{line} = \sum_{(i,j) \in \mathcal{L}} \left( \rho \parallel {\mathbf{r}_{\mathcal{L}}}\left(\mathbf{z}_{\mathcal{L}_{j}}^{c_{i}}, \mathcal{X}\right)\parallel_{\sum_{\mathcal{L}_{j}}^{c_{i}}}^{2} \right)  \\
 \rho(s)=\begin{cases}s & s\leq1 \\ 2\sqrt{s}-1 & s > 1\end{cases}
\end{cases}
\label{e:objectfunction}
\end{equation}
where $\rho(s)$ denotes the Huber norm \cite{huber}, which is used to suppress outliers. \par
Finally, with the objective function, the optimal state vector $\mathcal{X}$ can be found by iteratively updating from an initial guess $\mathcal{X}_{0}$:
$
\mathcal{X}_{t+1} = \mathcal{X}_{t} \oplus \delta\mathcal{X},
$
where $t$ denotes the $t$-th iteration, and $\delta$ denotes the increment. In particular, we introduce how to update $\mathbf{e}_{line}$. Based on Equation (\ref{e:qrcomposition})-(\ref{e:orthonormal}), $\mathbf{e}_{line}$ can be minimized by iteratively updating a minimum four-parameter orthonormal representation:
\begin{equation}
\begin{cases}\mathbf{U}_{t+1} \leftarrow  \mathbf{U}_{t}(\mathbf{I}+[\delta\bm{\theta}]_{\times}) \\ \mathbf{W}_{t+1} \leftarrow \mathbf{W}_{t} \left(\mathbf{I}+\begin{bmatrix}0 & -\delta\theta \\ \delta\theta & 0 \end{bmatrix} \right)  \end{cases},
\end{equation}
where $\mathbf{I}$ denotes the corresponding identity matrix. \par 

\begin{table}[t]
\centering
\caption{RMSE ATE [m] comparison of VINS-Mono and PL-VINS }
\label{t:ate}
\setlength{\tabcolsep}{0.9mm}{%
\begin{tabular}{@{}lcccc@{}}
\toprule
\multicolumn{1}{c}{\multirow{2}{*}{Datasets}} & \multicolumn{2}{c}{w/o loop}            & \multicolumn{2}{c}{w/ loop}           \\ \cmidrule(l){2-5} 
\multicolumn{1}{c}{}                          & VINS-Mono      & PL-VINS                & VINS-Mono      & PL-VINS              \\ \midrule
MH-04-\textit{difficult} 	& 0.375          & \textbf{0.270}         & 0.220          & \textbf{0.202}       \\
MH-05-\textit{difficult}                               & 0.296          & \textbf{0.272}         & \textbf{0.242} & 0.252                \\
V1-02-\textit{medium}                                  & \textbf{0.095} & 0.105                  & \textbf{0.084} & 0.092                \\
V1-03-\textit{difficult}                               & 0.176          & \textbf{0.156}         & 0.170          & \textbf{0.152}       \\
V2-03-\textit{difficult}                               & 0.293          & \textbf{0.237}         & 0.280          & \textbf{0.182}       \\
Mean                                         & 0.247          & \textbf{0.208 ($\downarrow$16\%)} & 0.199          & \textbf{0.175 ($\downarrow$12\%)} \\ \bottomrule
\end{tabular}%
}
\begin{flushleft}
All statics are collected from real reproduction test. \textit{Difficult} and \textit{Medium} represent the difficulty level of the sequence. Note that we jump \textit{Easy} sequences. The bold figure represents the better result. Generally, the localization error of PL-SLAM is 16\% w/o loop or 12\% w/ loop less than that of VINS-Mono.   
\end{flushleft}
\end{table}

\begin{table*}[t]
\centering
\caption{RMSE RPE [m] or [deg] Comparison of PL-VIO, VINS-Mono and PL-VINS}
\label{t:accuracy}
\setlength{\tabcolsep}{3mm}{%
\begin{tabular}{@{}lcccccccccccc@{}}
\toprule
\multicolumn{1}{c}{\multirow{3}{*}{Datasets}} & \multicolumn{6}{c}{fixed time delta = 1s  ($\backsim$1000 pairs)}                                                               & \multicolumn{6}{c}{no fixed time delta (over 9000 pairs)}                                                              \\ \cmidrule(l){2-13} 
\multicolumn{1}{c}{}                          & \multicolumn{2}{c}{PL-VIO} & \multicolumn{2}{c}{VINS-Mono}            & \multicolumn{2}{c}{PL-VINS}               & \multicolumn{2}{c}{PL-VIO} & \multicolumn{2}{c}{VINS-Mono}            & \multicolumn{2}{c}{PL-VINS}               \\ \cmidrule(l){2-13} 
\multicolumn{1}{c}{}                          & Trans.           & Rot.    & Trans.         & Rot.                    & Trans.         & Rot.                     & Trans.       & Rot.        & Trans.         & Rot.                    & Trans.         & Rot.                     \\ \midrule
MH-04-\textit{difficult}                                & 0.268            & 3.198   & \textbf{0.263} & \textit{\textbf{3.150}} & 0.269          & 3.216                    & 2.266        & 4.829       & 2.269          & 5.045                   & \textbf{2.258} & \textit{\textbf{4.781}}  \\
MH-05-\textit{difficult}                               & \textbf{0.251}   & 2.917   & 0.253          & 2.902                   & 0.253          & \textit{\textbf{2.850}}  & 2.336        & 4.112       & 2.342          & 4.109                   & \textbf{2.213} & \textit{\textbf{4.060}}  \\
V1-02-\textit{medium}                                  & 0.150            & 5.242   & 0.149          & \textit{\textbf{5.200}} & \textbf{0.148} & 5.263                    & 0.361        & 5.300       & \textbf{0.355} & \textit{\textbf{5.259}} & 0.365          & 5.317                    \\
V1-03-\textit{difficult}                               & 0.050            & 1.884   & 0.057          & 5.595                   & \textbf{0.046} & \textit{\textbf{1.728}}  & 0.270        & 2.830       & 0.314          & 4.573                   & \textbf{0.236} & \textit{\textbf{2.741}}  \\
V2-03-\textit{difficult}                               & 0.427            & 19.740  & 0.432          & 19.845                  & \textbf{0.420} & \textit{\textbf{19.702}} & 1.677        & 21.923      & 1.682          & 22.034                  & \textbf{1.640} & \textit{\textbf{21.742}} \\ \midrule
Best Count                                    & \multicolumn{2}{c}{1}      & \multicolumn{2}{c}{3}                    & \multicolumn{2}{c}{6}                     & \multicolumn{2}{c}{0}      & \multicolumn{2}{c}{2}                    & \multicolumn{2}{c}{9}                     \\ \bottomrule
\end{tabular}%
}
\begin{flushleft}
VINS-Mono* and PL-VINS* both run w/o loop as PL-VIO does not include it. All statics are collected from real reproduction test, RPE includes translation (Trans.) [m] and rotation (Rot.) [deg] error. Note that the evaluation result of RPE depends on the setting of the time delta. We examine two settings: fixed time delta = 1s and no fixed delta \cite{tum}. In our experiments, the former usually can find $\backsim$1000 pairs of poses while the latter can find over 9000 pairs. The last row (Best Count) counts the times of best value on the five challenging sequences. The bold figure represents the best Trans. value, the bold italic figure represents the best Rot. value. 
\end{flushleft}
\end{table*}

\begin{table*}[t]
\centering
\caption{Average Execution Time Comparison of VINS-Mono, PL-VIO and PL-VINS.}
\label{t:realtime}
\setlength{\tabcolsep}{3.8mm}{%
\begin{tabular}{@{}clcccccc@{}}
\toprule
\multirow{2}{*}{Threads} & \multicolumn{1}{c}{\multirow{2}{*}{Modules}} & \multicolumn{3}{c}{Times (ms)}           & \multicolumn{3}{c}{Rate (Hz)}                                            \\ \cmidrule(l){3-8} 
                         & \multicolumn{1}{c}{}                         & VINS-Mono & PL-VIO   & PL-VINS           & VINS-Mono           & PL-VIO                      & PL-VINS             \\ \midrule
1                        & Point Detection and Tracking                 & 15        & 15       & 15                & \multirow{3}{*}{25} & \multirow{3}{*}{\textbf{5}} & \multirow{3}{*}{10} \\
\textbf{}                & Line Detection                               & $\times$         & 70 (LSD) & 20 (Modified LSD) &                     &                             &                     \\
\textbf{}                & Line Tracking                                & $\times$         & 15       & 12                &                     &                             &                     \\ \midrule
2                        & Local VIO                    & 42        & 48       & 46                & \textbf{10}         & 10                          & \textbf{10}         \\ \midrule
3                        & Loop closure                                 & 200       &$\times$       &  200                &                     &                             &                     \\ \bottomrule
\end{tabular}%
}
\begin{flushleft}
$\times$ means the system does not implement it. The \textit{loop closure} thread optimizes pose in back-end for non-real time. \\ The bold figure represents the run rate. VINS-Mono and PL-VINS (ours) can run at 10Hz, PL-VIO 5hz.  
\end{flushleft}
\end{table*}

\section{EXPERIMENTS}
In this section, we evaluate the performance of PL-VINS in terms of localization accuracy and real-time performance on the benchmark dataset EuRoc \cite{euroc}. PL-VINS was implemented based on Ubuntu 18.04 with ROS Melodic, and all experiments were performed on a low-power Intel Core i7-10710U CPU @1.10 GHz.  

\subsection{Accuracy Comparison}
In this subsection, we test the localization accuracy of our PL-VINS method, which is evaluated by the root mean square error (RMSE) of the absolute trajectory error (ATE) and the relative pose error (RPE) \cite{euroc}. \par 
\textbf{ATE comparison}. Considering PL-SLAM is developed based on VINS-Mono, we first compare them on five challenging sequences of EuRoc. Table \ref{t:ate} provides an ATE comparison where the bold figure represents the better result. From this table, we can conclude that:
\begin{itemize}
	\item PL-VINS yields better localization accuracy except for V1-02-\textit{medium}. The ATE of our method can be as much as $\backsim$28\% (0.375$\rightarrow$0.270) less than that of VINS-Mono on the MH-04-\textit{difficult} sequence.
	\item \textit{loop closure} (loop) is a necessary step to eliminate accumulative error, which works on the all five sequences. Taking VINS-Mono as an example, 0.375 is reduced to 0.220 on the MH-04-\textit{difficult} sequence.
	\item To sum up from the last row, PL-SLAM is the better method as the localization error of PL-VINS is 16\% w/o loop or 12\% error w/ loop less than that of VINS-Mono, which demonstrates that line features can improve the accuracy of point-based VINS method.
\end{itemize}	
\par 
Fig. 4 provides an intuitive comparison in terms of 3D motion trajectory. Fig. \ref{f:ros} provides a visible example on the MH-04-\textit{Difficult} sequence, in which PL-VINS additionally rebuilds space lines. 
\par

\textbf{RPE comparison}. Table \ref{t:accuracy} provides a comparison of PL-VIO \cite{plvio}, VINS-Mono and PL-VINS, where the best value is bold. Note that RPE includes translational and rotational error, and its evaluation result depends on the setting of time delta \cite{tum}. We examine two settings: fixed time delta = 1s and no time fixed delta. The former usually can find $\backsim$1000 pairs of poses, and the latter can find over 9000 pairs in our experiments. In terms of the count result in the last row of this table, PL-VINS obtains 15 (6+9) best results while VINS-Mono 5 (3+2) and PL-VIO 1 (1+0). Therefore, PL-VINS yields better performance than the other two methods in terms of the RPE evaluation. Compared with PL-VIO, PL-VINS establishes more accurate line feature correspondence by the modified LSD algorithm in line feature extraction and the angular threshold in inlier refinement, obtaining better accuracy in terms of the RPE evaluation.


\subsection{Real-Time Analysis}
Table \ref{t:realtime} provides a comparison of average execution time of the three methods on the MH-04-\textit{difficult} sequence. Thread 1, 2 and 3 represent \textit{measurement preprocessing}, \textit{local VIO}, and \textit{loop closure}, respectively. ``Rate'' denotes the pose update frequency whose value is determined by the highest execution time in the three threads. We follow the real-time criterion in VINS-Mono which is that if the system can run at over 10 Hz pose update frequency while providing accurate localization information as output. \par
In our experiments, VINS-Mono and PL-VINS are both able to run at 10 Hz while PL-VIO fails. Note that VINS-Mono needs at most $\backsim$42 ms in Thread 2 while PL-VINS $\backsim$46 ms and PL-VIO $\backsim$100ms in Thread 1. Compared with PL-VIO, PL-VINS is more efficient by using the modified LSD algorithm.

\section{CONCLUSIONS}
This paper presents PL-VINS which is the first real-time optimization-based monocular VINS method with point and line features. In which, a modified LSD algorithm is presented for the pose estimation problem by studying a hidden parameter tuning and length rejection strategy. We argue that the modified LSD can be used for any other works related to pose estimation from line correspondences. In addition, we efficiently leverage line feature constraints in the optimization-based sliding window. As a result, PL-VINS can produce more accurate localization than the SOTA VINS-Mono \cite{vins} at the same pose update rate.\par 
It should be pointed out that line features still have not been exploited fully in this work. At least, the line feature constraints were not used for the \textit{loop closure} thread. Further, we observe that current works all perform frame-to-frame line feature matching between the last and the current frame, which may lead to the question why many line features in the previous several frames before the last frame are ignored although they can be probably observed by the current frame. Frame-to-model strategy may be one answer to this question, where the VINS system builds and maintains a local map model consisting of space lines, and then line correspondences are established between the current frame and the map model in order to make use of the ignored line features. This strategy is similar to the popular point-based frame-to-model strategy \cite{orbslam3}. \par 
For the future work, in addition to the above-mentioned questions, we also plan to extend the monocular vision to a stereo, and run experiments in more challenging and realistic environments, such as large-scale, low-texture, and low-light indoor scene to further study the utility of line feature constraints. 


%



\end{document}